\documentclass[10pt, conference]{IEEEtran}

\usepackage[cmex10]{amsmath}

\ifCLASSINFOpdf
   \usepackage[pdftex]{graphicx}
\else
   \usepackage[dvips]{graphicx}
\fi

\usepackage[caption=false,font=footnotesize]{subfig}

\usepackage{multirow}
\usepackage{multicol}
\usepackage{array}
\usepackage{cite}

\usepackage{wrapfig}   
\usepackage{textcomp}  

\usepackage{atbegshi,picture} 
\usepackage{eso-pic, rotating, graphicx}
\usepackage{url}

\usepackage[utf8]{inputenc}
\usepackage{array}

\usepackage{tikz}   
\usetikzlibrary{patterns,arrows,decorations.pathreplacing,fit,shapes}

\begin{document}

\title{SOM-Guided Evolutionary Search for Solving MinMax Multiple-TSP}

%
%
%
%

\author{\IEEEauthorblockN{Vlad-Ioan Lupoaie\IEEEauthorrefmark{1},
		Ivona-Alexandra Chili\IEEEauthorrefmark{2}, Mihaela Elena Breaban\IEEEauthorrefmark{3} and
		Madalina Raschip\IEEEauthorrefmark{4}}
	\IEEEauthorblockA{
	\IEEEauthorrefmark{1}\IEEEauthorrefmark{2}\IEEEauthorrefmark{3}\IEEEauthorrefmark{4}Faculty of Computer Science,
		Alexandru Ioan Cuza University of Iaşi,
		Romania\\
			\IEEEauthorrefmark{1}\IEEEauthorrefmark{2}Bitdefender Laboratory, Iaşi, Romania\\
		Email: \IEEEauthorrefmark{1}vlad.lupoaie@gmail.com,
		\IEEEauthorrefmark{2}ivona.chili@gmail.com,
		\IEEEauthorrefmark{3}pmihaela@info.uaic.ro,
		\IEEEauthorrefmark{4}mionita@info.uaic.ro}}

\maketitle

\begin{abstract}
Multiple-TSP, also abbreviated in the literature as mTSP, is an extension of the Traveling Salesman Problem that lies at the core of many variants of the Vehicle Routing problem of great practical importance. The current paper develops and experiments with Self Organizing Maps, Evolutionary Algorithms and Ant Colony Systems to tackle the MinMax formulation of the Single-Depot Multiple-TSP. Hybridization between the neural network approach and the two meta-heuristics shows to bring significant improvements, outperforming results reported in the literature on a set of problem instances taken from TSPLIB.
\end{abstract}


\section{Introduction}
Multiple-TSP is an extension of the well-known Traveling Salesman Problem where a number of $m$ salesmen are used to serve a set of $n$ locations/cities, restricted to the original constraint that each location must be visited only once. The agents/salesmen share a single depot - case known as Single-Depot Multiple-TSP, or use multiple depots - variant known as Multiple-Depot Multiple-TSP. Regarding the objective to be optimized, two distinct formulations generally exist: one that minimizes the total cost of visiting all the locations - which sums-up the costs of the individual tours of the $m$ agents (objective called \emph{MinSum}), and one that minimizes the cost of the longest tour (objective called \emph{MinMax}). 

The current paper deals with Single-depot Multiple-TSP. Both its \emph{MinSum} and \emph{MinMax} variants can be formalized as integer linear programming optimization problems \cite{ICTAI2015}. In previous studies we argue about the intrinsic bi-objective nature of Multiple-TSP: in practice, it is desirable to obtain minimum cost with regard to the total traveled distance while at the same time having a balanced workload (the tours of the $m$ salesmen should have approximately equal costs). From this point of view, the \emph{MinSum} formulation is ill-posed, leading to degenerated solutions where one of the agents takes-over most of the work. To deal with this issue, new constraints may be imposed in the MinSum formulation, restricting the number of cities to be visited to an interval - version called Bounded Multiple-TSP; its formulation as an integer optimization problem and approaches based on Ant Systems can be found in our previous work \cite{HAIS2015}.

By minimizing the maximum tour length of the solution, the MinMax Single-Depot Multiple-TSP intrinsically performs total cost minimization while achieving balanced tours. This is the problem we tackle in this study by means of three different paradigms and their hybrids. Section \ref{sec:related} reviews approaches reported in the literature for this specific version of the problem. In section \ref{sec:algs} we describe in detail the algorithms we propose while in the experimental section \ref{sec:experiments} we analyze comparatively their performance and highlight the gains achieved by their hybrids. Section \ref{sec:conclusion} concludes the paper.

\section{Related work} \label{sec:related}
While the literature abounds in studies that approach TSP, Multiple-TSP is much less investigated, despite the fact that it stands as basis for many variants of routing and scheduling problems. Moreover, some studies reduce Multiple-TSP to TSP by duplicating the depot in accordance with the number of salesmen, and apply TSP-specific algorithms, approach which has been shown to actually produce degenerated instances that are more difficult to solve, compared to an ordinary TSP instance \cite{Bektas2006}. Multiple-TSP has been approached in all its variants, but this section reviews only the \emph{MinMax Single-depot Multiple-TSP} formulation which actually received much less attention in the literature.   

Tabu search and two exact algorithms described in \cite{Franca95} are among the first approaches used for the MinMax Multiple-TSP. Similar results for symmetric and asymmetric instances were obtained by the authors. Tabu Search is also compared to a chaotic search method in \cite{Matsuura14}.

A clustering approach is proposed in \cite{Chandran06} in order to balance the workloads amongst salesmen; after the clusters are created, the nearest neighbour heuristic is used to build the route within each cluster. 

A few approaches based on Evolutionary Computing algorithms were reported for the Min-Max Multiple-TSP. In \cite{Wang17} a memetic algorithm based on a sequential variable neighborhood descent procedure is proposed. The used neighborhood considers moving the cities from one tour to another unidirectionally. A team ACO algorithm is proposed in \cite{Vallivaara08}. A team of ants construct the routes in parallel. In order to distribute the workload, the member of the team with the minimum route length is chosen to make a move.

In our previous work clustering was employed in conjunction with an ACO algorithm \cite{EVOLVE2015}; both crisp and fuzzy partitions were used to guide the ants while building the tours. In \cite{ICTAI2015} we comparatively evaluated three multi-objective ACO
based algorithms against a single-objective ACO algorithm guided by
the MinMax objective, from a bi-objective perspective counting for the total cost and the minimum imbalance measured as the difference between the costs of the longest and the shortest tour. Using Pareto dominance concepts and dedicated metric, the experimental analysis has shown the superiority of the MinMax approach,
that although being the simplest one, achieved good tradeoff
solutions that are diverse and in proximity of the reference
Pareto front.

The problem was also addressed by neural networks. Self organizing maps were used in \cite{Modares99} to tackle the MinMax variant of Multiple-TSP and vehicle routing.

\section{The algorithms investigated} \label{sec:algs}
The aim of the current paper is to devise a hybrid algorithm for solving the MinMax Single-depot Multiple-TSP which combines different key concepts from state-of-the art methods. Self Organizing Maps, which are Machine Learning methods that aim at generating topological mappings that reflect semantic relationships in data, can be used to derive an arrangement/ordering of the cities that exhibit shortest-path properties without explicitly minimizing the cost of the tours. On the other hand, Evolutionary Computation techniques like Evolutionary Algorithms and Ant Colony Optimization are general-purpose optimization methods that perform the search towards the optimal solution guided by the objective function to be optimized. We analyze the performance of each method for the problem we tackle and try to obtain some benefits from combining the different approaches. In the following, each method we implement is presented in detail, framed in the literature context from where we draw its inspiration.

\subsection{SOM for MinMax multiple-TSP} \label{sec:SOM_algorithm}
Neural networks are algorithms used primarily to solve classification problems \cite{gavrilut1, gavrilut2}. Self Organizing Maps (SOM), also called Kohonen maps, are a special type of neural networks used in the field of Unsupervised Machine Learning. Their use to address shortest-path problems date back to '88 in the context of TSP \cite{SOM88a,SOM88b}. A first extension  to MinMax multiple-TSP is presented in \cite{Modares99} and direct applications for multirobot path planning in \cite{Faigl16}. Subsequent papers present extensions that incorporate various constraints for VRP.

SOM is usually presented as a neural network consisting of two layers: the input layer where the data is fed and the output layer which consists of a number of neurons arranged on a regular surface. When solving TSP, the data to be fed consists of the set of cities which are expressed as 2-dimensional numerical vectors (the coordinates of the cities) and in the output layer the neurons use a ring topology. Their weight vectors are initialized in the same 2-dimensional space arranged on a circle with the center initialized to be the center of the set of cities. The standard SOM training algorithm is used: in an iterative process each city is given as input and a competition among output neurons follows which designates as winner the neuron at the smallest distance. Then, implementing the cooperation concept, the winning neuron but also its neighbors on the ring are updated by changing their weights towards the coordinates of the input city. After a number of iterations, the algorithm converges and the TSP solution is built by traversing the cities in the order of their winning neurons, as initialized on the circle.

\begin{figure}[h]
	\centering
	\includegraphics[width = 4.2cm]{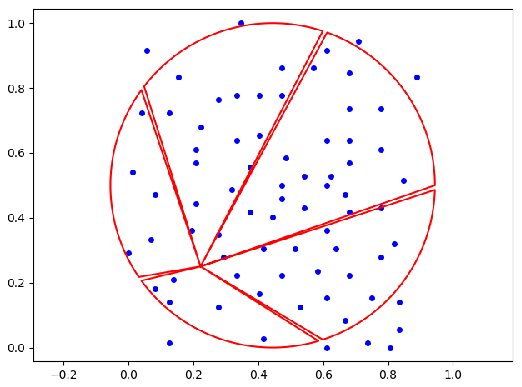}
		\includegraphics[width = 4.2cm]{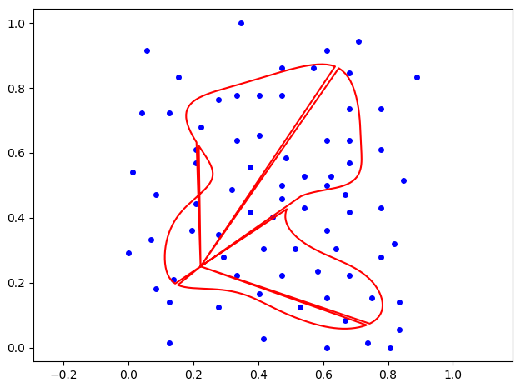}
			\includegraphics[width = 4.2cm]{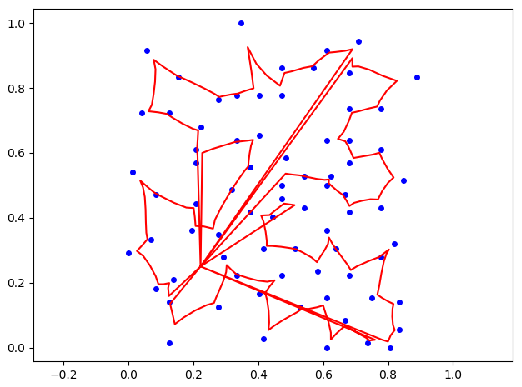}
				\includegraphics[width = 4.2cm]{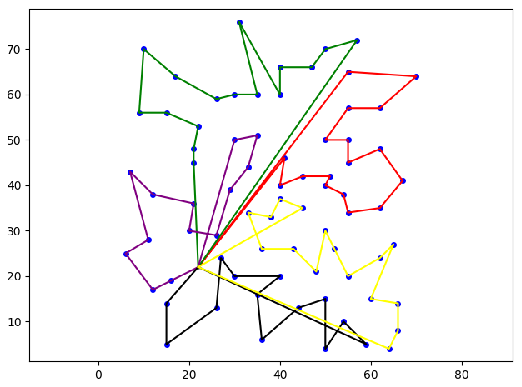}
	\caption{SOM for Multiple-TSP on instance eil76 with 5 salesmen: initialization, intermediate iteration, final iteration and final solution. Blue points correspond to cities, red curves correspond to the trajectory given by traversing the neurons using the ring topology}
	\label{fig:SOM}
\end{figure}

In case of Multiple-TSP, the topology of the output layer must be modified such that instead of a single route, a number of $m$ routes must be generated. In \cite{Modares99}, a number of $m$ circles are generated, each circle initializing one of its neurons to the coordinates of the depot. In contrast, our initialization is similar to the TSP scenario using only one circle centered in the city space but also adding one more neuron which is initialized to the coordinates of the depot. The topology is thus slightly changed when compared to TSP: not all the neurons have as nearest neighbors the neurons situated at their immediate left and right on the circle, but for a number of $m$ times the circle is discontinued to interleave the neuron corresponding to the depot. Figure \ref{fig:SOM} illustrates the initialization step, an intermediate and the final state of the training process; the solution is represented in the fourth graph.

The training process for Multiple-TSP can be summarized as follows. At each iteration a city $x$ is drawn randomly and the winning neuron, denoted by $w^*$ is chosen to be the one at the smallest Euclidean distance. Then, the radius of the neighbourhood of the winning neuron is computed. The value of the radius is initially set to the number of neurons but diminishes at each iteration, using exponential decay shown in equation \ref{eq:radius}:

\begin{equation}\label{eq:radius}
\sigma(t)=\sigma_0 e^{-t/\lambda_r}. 
\end{equation}
where $t$ is the current iteration, $\sigma_0$ denotes the initial value of the radius and $\lambda_r$ is a decay constant equal to the total number of iterations $k$ divided by $\log(\sigma_0)$. 

Any nodes within winner neurons' neighbourhood, excepting the node corresponding to the depot which is never changed,  have their weight altered according to formulas \ref{eq:SOM} and \ref{eq:dist}:
\begin{equation}\label{eq:SOM}
w_i=w_i+\alpha n(w^*,i)(x-w_i)  
\end{equation}

\begin{equation}\label{eq:dist}
n(w^*, i)=e^{\frac{-dist(w^*, i)^2}{2*(\sigma(t)/10)^2}}. 
\end{equation}
where $dist$ computes the distance on the ring and not in the weight space.

The learning rate $\alpha$ also decays at each iteration using the exponential decay function \ref{eq:lr}:
\begin{equation}\label{eq:lr}
\alpha(t)=\alpha_0e^{-t/\lambda_l}.
\end{equation}
where $t$ is the current iteration, $\alpha_0$ denotes the initial learning rate and $\lambda_l = k / \log(\alpha_0 / \alpha_{min})$ where $\alpha_{min}$ is the minimum learning rate.

The final solution is built after the last iteration of the SOM: for each city we compute again the closest neuron and the tours are formed based on the relative order of these neurons on the initial ring topology.

\subsection{The evolutionary algorithm}
The evolutionary algorithm we use is a kind of Evolution Strategy that implements several mutation operators to create offsprings by altering a single individual (candidate solution), without resorting to crossover.

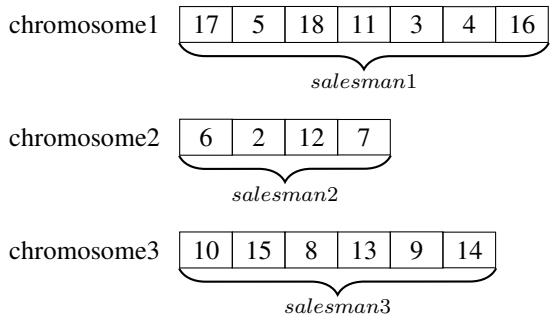
\begin{figure}[h]
\centering
\begin{tikzpicture}
    \node [left] at (1, -1) {chromosome1};
    \node (n11) [draw, minimum width=0.7cm] at (1.5, -1) {17};
    \node (n21) [draw, minimum width=0.7cm] at (2.2, -1) {5};
    \node (n31) [draw, minimum width=0.7cm] at (2.9, -1) {18};
    \node (n41) [draw, minimum width=0.7cm] at (3.6, -1) {11};
    \node (n51) [draw, minimum width=0.7cm] at (4.3, -1) {3};
    \node (n61) [draw, minimum width=0.7cm] at (5, -1) {4};
    \node (n71) [draw, minimum width=0.7cm] at (5.7, -1) {16};
    \draw [thick, black, decorate, decoration={brace, amplitude=10pt, mirror}, xshift=0.2pt, yshift=-0.2pt](n11.south west) -- (n71.south east) node[black, midway, yshift=-0.5cm] {\footnotesize $salesman1$};
    
    \node [left] at (1, -2.5) {chromosome2};
    \node (n12) [draw, minimum width=0.7cm] at (1.5, -2.5) {6};
    \node (n22) [draw, minimum width=0.7cm] at (2.2, -2.5) {2};
    \node (n32) [draw, minimum width=0.7cm] at (2.9, -2.5) {12};
    \node (n42) [draw, minimum width=0.7cm] at (3.6, -2.5) {7};
    \draw [thick, black, decorate, decoration={brace, amplitude=10pt, mirror}, xshift=0.2pt, yshift=-0.2pt](n12.south west) -- (n42.south east) node[black, midway, yshift=-0.5cm] {\footnotesize $salesman2$};
    
    \node [left] at (1, -4) {chromosome3};
    \node (n13) [draw, minimum width=0.7cm] at (1.5, -4) {10};
    \node (n23) [draw, minimum width=0.7cm] at (2.2, -4) {15};
    \node (n33) [draw, minimum width=0.7cm] at (2.9, -4) {8};
    \node (n43) [draw, minimum width=0.7cm] at (3.6, -4) {13};
    \node (n53) [draw, minimum width=0.7cm] at (4.3, -4) {9};
    \node (n63) [draw, minimum width=0.7cm] at (5, -4) {14};
    \draw [thick, black, decorate, decoration={brace, amplitude=10pt, mirror}, xshift=0.2pt, yshift=-0.2pt](n13.south west) -- (n63.south east) node[black, midway, yshift=-0.5cm] {\footnotesize $salesman3$};
\end{tikzpicture}
\caption{The multi-chromosome representation for 3 salesmen and 17 locations}
\label{fig:multichromosome}
\end{figure}

One popular representation proposed for the evolutionary algorithms for solving Multiple-TSP was the two-part chromosome representation \cite{Carter06}. This technique reduces the size of the search space by eliminating redundant solutions. However, because of the need to perform more complex mutations within an individual triggered by the necessity of balancing the tours, we have used instead the recently proposed representation, the multi-chromosome technique \cite{Kiraly11}, that showed good results for the MinSum variant of Multiple-TSP. Example of an individual modeled with the chosen technique can be seen in Figure \ref{fig:multichromosome}. The depot, denoted by the node on the first position (node number 1), should be redundantly placed at the beginning and end of each chromosome and it is therefore not stored in the representation.

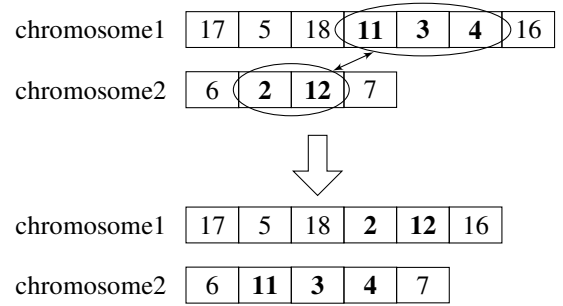
\begin{figure}[h]
\centering
\begin{tikzpicture}
    \node [left] at (1, -1) {chromosome1};
    \node (n11) [draw, minimum width=0.7cm] at (1.5, -1) {17};
    \node (n21) [draw, minimum width=0.7cm] at (2.2, -1) {5};
    \node (n31) [draw, minimum width=0.7cm] at (2.9, -1) {18};
    \node (n41) [draw, minimum width=0.7cm] at (3.6, -1) {\textbf{11}};
    \node (n51) [draw, minimum width=0.7cm] at (4.3, -1) {\textbf{3}};
    \node (n61) [draw, minimum width=0.7cm] at (5, -1) {\textbf{4}};
    \node (n71) [draw, minimum width=0.7cm] at (5.7, -1) {16};
    \node[draw=black, fit=(n41)(n51)(n61), inner sep=0pt, text width=47pt, ellipse] (first) {};
    
    \node [left] at (1, -1.8) {chromosome2};
    \node (n12) [draw, minimum width=0.7cm] at (1.5, -1.8) {6};
    \node (n22) [draw, minimum width=0.7cm] at (2.2, -1.8) {\textbf{2}};
    \node (n32) [draw, minimum width=0.7cm] at (2.9, -1.8) {\textbf{12}};
    \node (n42) [draw, minimum width=0.7cm] at (3.6, -1.8) {7};
    \node[draw=black, fit=(n22)(n32), inner sep=0pt, text width=31pt, ellipse] (second) {};
    
    \draw[latex'-latex']  (first) edge (second);
    
    \node[draw, single arrow, minimum height=8mm, minimum width=5mm, single arrow head extend=2mm, anchor=west, rotate=-90] at (2.9, -2.4) {};
    
    \node [left] at (1, -3.6) {chromosome1};
    \node (m11) [draw, minimum width=0.7cm] at (1.5, -3.6) {17};
    \node (m21) [draw, minimum width=0.7cm] at (2.2, -3.6) {5};
    \node (m31) [draw, minimum width=0.7cm] at (2.9, -3.6) {18};
    \node (m41) [draw, minimum width=0.7cm] at (3.6, -3.6) {\textbf{2}};
    \node (m51) [draw, minimum width=0.7cm] at (4.3, -3.6) {\textbf{12}};
    \node (m71) [draw, minimum width=0.7cm] at (5, -3.6) {16};
    
    \node [left] at (1, -4.4) {chromosome2};
    \node (m12) [draw, minimum width=0.7cm] at (1.5, -4.4) {6};
    \node (m22) [draw, minimum width=0.7cm] at (2.2, -4.4) {\textbf{11}};
    \node (m22) [draw, minimum width=0.7cm] at (2.9, -4.4) {\textbf{3}};
    \node (m22) [draw, minimum width=0.7cm] at (3.6, -4.4) {\textbf{4}};
    \node (m42) [draw, minimum width=0.7cm] at (4.3, -4.4) {7};
\end{tikzpicture}
\caption{Cross-tour gene sequence transposition}
\label{fig:crossover}
\end{figure}

This representation closely resembles the constraints of the problem, different salesmen being physically separated, thus eluding the inherent challenges that crossover operators have on single chromosome models. The crossover operator specific to genetic algorithms is transformed into a cross-tour mutation and the different individuals do not exchange any information between them. A simple but effective technique, the gene sequence transposition depicted in Figure \ref{fig:crossover}, is implemented in order to ensure the exchange of cities between different tours. Each chromosome is selected for cross-tour mutation with a probability of $p_x$.


Aiming at balanced tours in the case of MinMax mTSP, we propose a small modification in the selection of candidate pairs for cross-tour mutation: instead of randomly building pairs from the selected chromosomes, we introduce a chance $p_{sort}$ for those selected to be sorted according to their individual fitness and the pairs built picking the first candidate from one end and the other candidate from the other end (longest tour with shortest tour, second longest tour with second shortest tour etc.), thus giving better individuals a chance to cooperate with and improve worse ones.

For intra-tour mutations all three methods introduced in \cite{Kiraly11} were implemented. A short description of these is presented below.

\begin{figure}[h]
\centering
\begin{tikzpicture}
    \node (n11) [draw, minimum width=0.7cm] at (1.5, -1) {17};
    \node (n21) [draw, minimum width=0.7cm] at (2.2, -1) {5};
    \node (n31) [draw, minimum width=0.7cm] at (2.9, -1) {\textbf{18}};
    \node (n41) [draw, minimum width=0.7cm] at (3.6, -1) {\textbf{11}};
    \node (n51) [draw, minimum width=0.7cm] at (4.3, -1) {\textbf{3}};
    \node (n61) [draw, minimum width=0.7cm] at (5, -1) {\textbf{4}};
    \node (n71) [draw, minimum width=0.7cm] at (5.7, -1) {16};
    
    \draw[color=black] (2.55, -1) -- (2.55, -0.5);
    \draw[color=black] (5.35, -1) -- (5.35, -0.5);
    \draw[-latex, thick, color=black] (5.25, -0.6) to [bend right] (2.65, -0.6);
    
    \node[draw, single arrow, minimum height=8mm, minimum width=5mm, single arrow head extend=2mm, anchor=west, rotate=-90] at (3.62, -1.4) {};
    
    \node (m11) [draw, minimum width=0.7cm] at (1.5, -2.6) {17};
    \node (m21) [draw, minimum width=0.7cm] at (2.2, -2.6) {5};
    \node (m31) [draw, minimum width=0.7cm] at (2.9, -2.6) {\textbf{4}};
    \node (m41) [draw, minimum width=0.7cm] at (3.6, -2.6) {\textbf{3}};
    \node (m51) [draw, minimum width=0.7cm] at (4.3, -2.6) {\textbf{11}};
    \node (m61) [draw, minimum width=0.7cm] at (5, -2.6) {\textbf{18}};
    \node (m71) [draw, minimum width=0.7cm] at (5.7, -2.6) {16};

\end{tikzpicture}
\caption{In-tour gene sequence inversion}
\label{fig:inversion}
\end{figure}
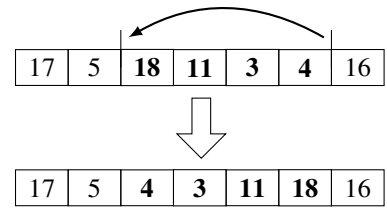

The in-tour gene sequence inversion operator depicted in Figure \ref{fig:inversion} randomly picks two cutting points with uniform probability $\frac{p_{si}}{k}$, where $p_{si}$ is the probability of sequence inversion mutation and $k$ is the number of cities in the mutated chromosome, and reverses the order of cities between the chosen points.

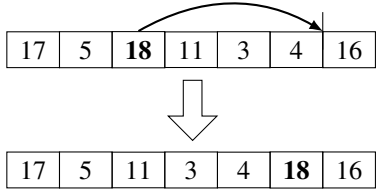
\begin{figure}[h]
\centering
\begin{tikzpicture}
    \node (n11) [draw, minimum width=0.7cm] at (1.5, -1) {17};
    \node (n21) [draw, minimum width=0.7cm] at (2.2, -1) {5};
    \node (n31) [draw, minimum width=0.7cm] at (2.9, -1) {\textbf{18}};
    \node (n41) [draw, minimum width=0.7cm] at (3.6, -1) {11};
    \node (n51) [draw, minimum width=0.7cm] at (4.3, -1) {3};
    \node (n61) [draw, minimum width=0.7cm] at (5, -1) {4};
    \node (n71) [draw, minimum width=0.7cm] at (5.7, -1) {16};
    
    \draw[color=black] (5.35, -1) -- (5.35, -0.5);
    \draw[-latex, thick, color=black] (n31.north) to [bend left] (n61.north east);
    
    \node[draw, single arrow, minimum height=8mm, minimum width=5mm, single arrow head extend=2mm, anchor=west, rotate=-90] at (3.62, -1.4) {};
    
    \node (m11) [draw, minimum width=0.7cm] at (1.5, -2.6) {17};
    \node (m21) [draw, minimum width=0.7cm] at (2.2, -2.6) {5};
    \node (m31) [draw, minimum width=0.7cm] at (2.9, -2.6) {11};
    \node (m41) [draw, minimum width=0.7cm] at (3.6, -2.6) {3};
    \node (m51) [draw, minimum width=0.7cm] at (4.3, -2.6) {4};
    \node (m61) [draw, minimum width=0.7cm] at (5, -2.6) {\textbf{18}};
    \node (m71) [draw, minimum width=0.7cm] at (5.7, -2.6) {16};

\end{tikzpicture}
\caption{In-tour gene insertion}
\label{fig:insertion}
\end{figure}

The in-tour gene insertion operator detailed in Figure \ref{fig:insertion} randomly picks a node and a cutting point with uniform probability $\frac{p_{in}}{k}$, where $p_{in}$ is the probability of insertion mutation and $k$ is the number of cities in the mutated chromosome, and moves the chosen city to the designated position.

\begin{figure}[h]
\centering
\begin{tikzpicture}
    \node (n11) [draw, minimum width=0.7cm] at (1.5, -1) {17};
    \node (n21) [draw, minimum width=0.7cm] at (2.2, -1) {5};
    \node (n31) [draw, minimum width=0.7cm] at (2.9, -1) {\textbf{18}};
    \node (n41) [draw, minimum width=0.7cm] at (3.6, -1) {11};
    \node (n51) [draw, minimum width=0.7cm] at (4.3, -1) {3};
    \node (n61) [draw, minimum width=0.7cm] at (5, -1) {\textbf{4}};
    \node (n71) [draw, minimum width=0.7cm] at (5.7, -1) {16};
    
    \draw[latex'-latex', thick, color=black] (n31.north) to [bend left] (n61.north);
    
    \node[draw, single arrow, minimum height=8mm, minimum width=5mm, single arrow head extend=2mm, anchor=west, rotate=-90] at (3.62, -1.4) {};
    
    \node (m11) [draw, minimum width=0.7cm] at (1.5, -2.6) {17};
    \node (m21) [draw, minimum width=0.7cm] at (2.2, -2.6) {5};
    \node (m31) [draw, minimum width=0.7cm] at (2.9, -2.6) {\textbf{4}};
    \node (m41) [draw, minimum width=0.7cm] at (3.6, -2.6) {11};
    \node (m51) [draw, minimum width=0.7cm] at (4.3, -2.6) {3};
    \node (m61) [draw, minimum width=0.7cm] at (5, -2.6) {\textbf{18}};
    \node (m71) [draw, minimum width=0.7cm] at (5.7, -2.6) {16};

\end{tikzpicture}
\caption{In-tour gene transposition}
\label{fig:transposition}
\end{figure}
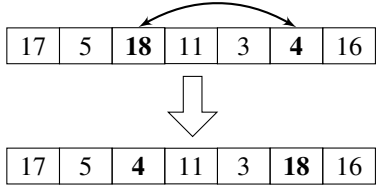

The in-tour gene transposition operator represented in Figure \ref{fig:transposition} acts like a double in-tour gene insertion, randomly picking two nodes with uniform probability $\frac{p_{tr}}{k}$, where $p_{tr}$ is the probability of transposition mutation and $k$ is the number of cities in the mutated chromosome, and swaps the cities between them.

The value of each mutation probability is derived from the global in-tour mutation probability $p_m$ as shown in formula \ref{eq:mutationprob}:

\begin{equation}\label{eq:mutationprob}
p_{si} = p_{in} = p_{tr} = \frac{p_m}{3}. 
\end{equation}

The selection of individuals from current generation to the next one uses the wheel of fortune method, which gives each individual a chance of survival proportional to its fitness. In order to preserve the best individuals and encourage progress we employ a global elitism strategy adding back to the population after each selection phase a percentage $e$ of historically best individuals.

To further improve on obtained solutions our approach implements the well known 2-opt local search heuristic \cite{Croes58}, which is run once every $f_{2opt}$ generations on all the tours of all individuals. A lower value of $f_{2opt}$ will degrade the time performance of the algorithm, as the mentioned local search heuristic is computationally expensive. Section \ref{sec:experiments} displays the results we have obtained with and without the 2-opt heuristic.

\subsection{The ACO approach}
Although ACO algorithms are state-of-the-art methods for solving TSP, in the case of Multiple-TSP only a few approaches exist. Generally, ACO builds the solution sequentially: for TSP, at each step a city is sampled in accordance with a probability distribution, while for Multiple-TSP, at each step both the  salesman and the city must be chosen. In our previous work \cite{HAIS2015, ICTAI2015, EVOLVE2015} we experimented with several ACO algorithms for solving Multiple-TSP in a bi-objective manner with the aim to minimize the total cost and the degree of imbalance of the tours within the solution. Among the algorithms we proposed, we choose to use here the version called \emph{g-MinMaxACS}, which showed to outperform more complex schemes in terms of multi-objective evaluation criteria. We will summarize here the algorithm, but a more elaborated description can be found in \cite{EVOLVE2015}.

g-MinMaxACS is a variation of the Ant Colony System: instead of using a single ant to construct one tour as in TSP, a set of $m$ ants is used to generate a complete solution to the Multiple-TSP problem. Initially, all ants start their tours from the depot and the salesman to visit the next city is selected at random. The selected salesman chooses the next city it will visit according to the transition rule from the standard ACS algorithm shown in formula \ref{eq:ACO_selection}.
\begin{equation} \label{eq:ACO_selection}
 s = \left\{
\begin{array}{l l}
\arg \max_{u \in C}{\tau(r,u)\cdot \eta^{\beta}(r,u)}, & \text{if $q \leq q_0$}\ \ \\\
S, & \text{otherwise}\ \ \ 
\end{array} \right.
\end{equation}

where $\tau(r,s)$ is the pheromone trail associated to edge $(r,s)$, $\eta(r,s)$ denotes the heuristic information corresponding to edge $(r,s)$ taken as the inverse of the cost measure (distance), $\beta$ is a parameter that reflects the relative importance of pheromone vs. distance, $q$ is a random number chosen in the range [0,1] and $q_0 \in [0,1]$ is a parameter.

$S$ is a random variable with the probability distribution given by:
\begin{equation}
 p(r,s) = \left\{
\begin{array}{l l}
\frac{\tau(r,s)\cdot \eta^{\beta}(r,s)}{\sum_{u\in C}\tau(r,u)\cdot \eta^{\beta}(r,u)}, & \text{if $s \in C$} \\
0, & \text{otherwise}
\end{array} \right.
\label{eq:p_node_rs}
\end{equation}
which defines the probability with which an ant chooses to move next to node $s$, while being in node $r$.

When an ant crosses an edge, the local pheromone update takes place and the pheromone level on the visited edge is updated like in the standard ACS, regardless of which salesman traverses it, using formula \ref{eq:ACO_local}. 

\begin{equation}\label{eq:ACO_local}
\tau(r,s)=(1-\rho)\cdot \tau(r,s)+\rho\cdot \Delta \tau(r,s)
\end{equation}
where $\rho \in (0,1)$ is a local pheromone decay parameter and
$\Delta \tau(r,s) = \tau_0$, where $\tau_0$ is the initial pheromone level, and is defined as:
\begin{equation}
\tau_0=(n\cdot L_{NN})^{-1}
\label{eq:initial_pheromone}
\end{equation}
which denotes a quantity set in the first iteration of the algorithm by building in a greedy manner based on a nearest-neighbor heuristic a tour and computing its length denoted by $L_{NN}$, and $n$ is the number of cities in the TSP instance.

This process continues until there are no remaining unvisited cities, meaning that a complete candidate solution was built. The best solution is selected to be the one with the smallest cost for its longest tour (in accordance with the MinMax objective), which corresponds to the global best solution. After all ants finish to construct their tours, a global pheromone update is applied to the edges used by the best ant. The amount
of deposited pheromone is given by the cost of the longest tour of the global best solution. The pheromone is updated according to formula \ref{eq:ACO_global}:

\begin{equation}\label{eq:ACO_global}
\tau(r,s)=(1-\alpha)\cdot \tau(r,s)+\alpha\cdot \Delta \tau(r,s)
\end{equation}
where
\[ \Delta \tau(r,s) = \left\{
\begin{array}{l l}
(L_{gb})^{-1} , & \text{if $(r,s) \in $ global-best-tour}\\
0, & \text{otherwise}
\end{array} \right.\]
where $\alpha \in (0,1)$ is the pheromone decay parameter and $L_{gb}$ designates the length of the global best tour found so far.

\subsection{Hybridization}
The solution generated by SOM for Multiple-TSP is dependent on initialization, i.e. on the locations used to interleave the depot neuron with neurons on the circle (it must be interleaved for a number of $m$ times, at equal distances in order to generate $m$ balanced tours in a run, as shown in Figure \ref{fig:SOM}). Therefore, a population of SOM solutions is generated by rotating the positions where the depot neuron is connected with the neurons on the circle. This initial solutions are further used to proceed with the search phase realized with the Evolutionary and ACO algorithms. In case of the evolutionary algorithm, the entire population is generated by running SOM with distinct initializations, while in case of g-MinMaxACS the pheromone trails are initialized using the tours generated by SOM in several runs. The formula used to initialize the pheromone trails is given below
\begin{equation}
\tau_0=\sum_{i=1..N}(n\cdot L_{SOM_i})^{-1}
\label{}
\end{equation}
where $N$ is the number of solutions obtained with SOM using $N$ different initializations and $L_{SOM_i}$ is the length of the solution given by SOM in run $i$.

\section{Experiments} \label{sec:experiments}
The experimental analysis is focused on analyzing the comparative performance of SOM, the evolutionary and the ACS-based algorithms and on investigating the boost in performance that can be obtained when SOM is used in conjunction with the two meta-heuristics.

\subsection{Problem instances} \label{sec:instances}
In order to evaluate the quality of the solutions obtained by the proposed approaches, the following TSP problem instances from the TSPLIB library were selected: eil51, berlin52, eil76 and rat99. For each problem instance, we considered 4 different values (2, 3, 5 and 7) for the number of salesmen. Consequentially, we executed the algorithms on all 16 instances. The description of these instances and the solutions obtained by the CPLEX solver for the MinMax formulation are available on the multiple-TSP page \footnote{\url{https://profs.info.uaic.ro/~mtsplib/MinMaxMTSP/}}. The MinMax variant is more difficult to solve than the MinSum multiple-TSP. For a small number of problems the optimal length of the longest tour is identified by the solver, for the others, a lower and an upper bound is found due to the time limits imposed (of several days), as specified on the resource web page.

\subsection{Parameter settings} \label{sec:params}
For all the algorithms investigated, the numerical parameters were empirically optimized. We report here the values that were chosen and produce the results illustrated in the next section.

\begin{figure}[ht]
	\centering
	\includegraphics[width = 8cm]{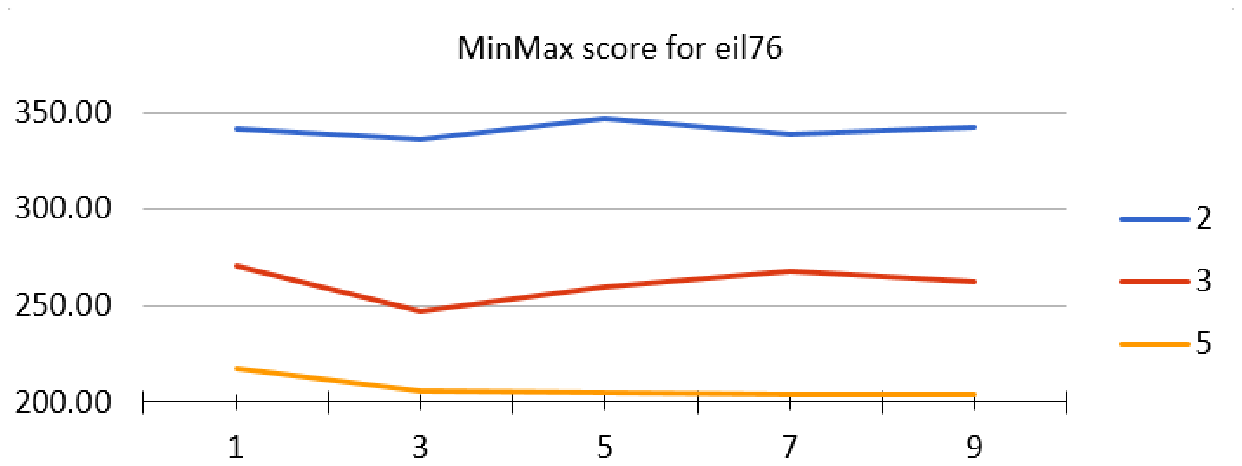}
		\includegraphics[width = 8cm]{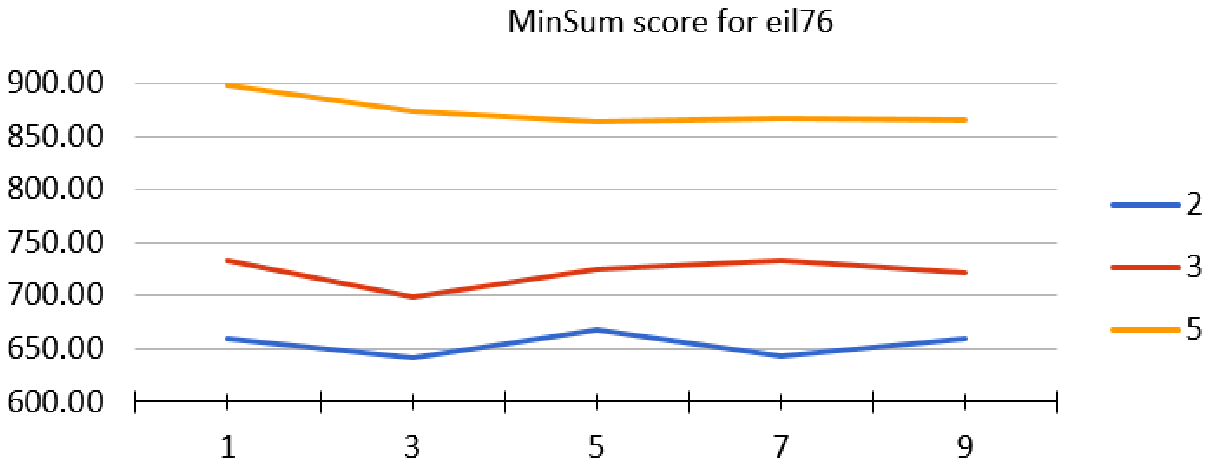}

	\caption{Results obtained with SOM using a number of neurons $d$ times the number of cities, with $d$ varying from 1 to 9. The first chart illustrates the MinMax score while the second the MinSum score computed for the same solutions. The three lines correspond to different number of salesmen}
	\label{fig:SOM-neurons}
\end{figure}

The numerical parameters for SOM were set as follows: learning rate $\alpha = 0.6$, minimum learning rate $\alpha_{min} = 0.01$, number of iterations $k = 5000$ which empirically showed to be sufficient to provide good initial solutions for the evolutionary algorithms. We studied the influence the number of neurons has on the quality of the solution. The number of neurons was set as a multiple of the number of cities, $d * n$ (where $n$ is the number of cities). Figure \ref{fig:SOM-neurons} illustrates some results obtained when varying $d$ between 1 and 9 on $eil76$ problem instance. Because the computational effort of SOM increases with the number of neurons, we are interested in the smallest number of neurons that produce good results, therefore a value $d=3$ was chosen.

For the evolutionary algorithm the parameters were set as follows: the population size was set to $100$, the cross-tour mutation probability $p_x = 0.4$ and the probability of sorting before pair matching is $p_{sort} = 0.1$, global in-tour mutation rate $p_m = 0.1$, percentage of historically best individuals re-added by the elitist strategy $e = 0.25$ and 2-opt local search heuristic is run once every $f_{2opt} = 10$ generations.

For g-MinMaxACS we used the parameter settings reported in the previous study \cite{EVOLVE2015}:  $q_0 = 0.9$, $\alpha = 0.1$, $\rho = 0.1$, $\beta = 2.0$.

For the evolutionary and ant algorithms we imposed the same number of fitness evaluations decided based on inspecting the convergence of both algorithms in several runs. Thus, a number of 250000 fitness evaluations was decided for all the experiments. Fitness evaluations of the 2-opt heuristic in the case of SOM-EA-2opt method are not counted towards the before mentioned total.

\subsection{Results} \label{sec:results}
The results for the proposed approaches are provided next. The performance of the following methods was analyzed: the Self Organizing Map algorithm (SOM), the Ant Colony g-MinMaxACS algorithm (ACO), the bybrid approach of SOM and g-MinMaxACS (SOM-ACO), the evolutionary algorithm (EA), the hybrid approach of SOM and EA (SOM-EA) and the hybridization between SOM, EA and 2-opt (SOM-EA-2opt).

The performance measure computed by all algorithms is the cost of the longest tour, the equivalent of the MinMax criterion. All algorithms, except for SOM, were executed for 50 runs and the average measure is reported. For the SOM algorithm we considered 300 runs.

\begin{figure}[ht]
	\centering
	\includegraphics[width = 9cm]{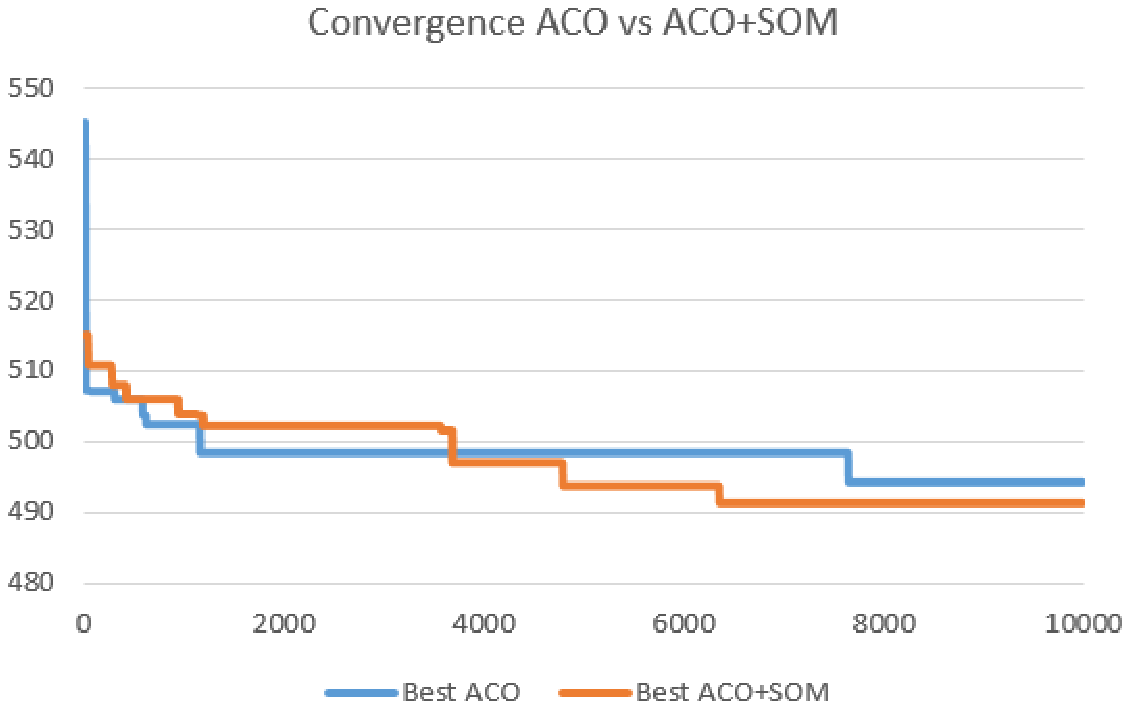}
	\includegraphics[width = 9cm]{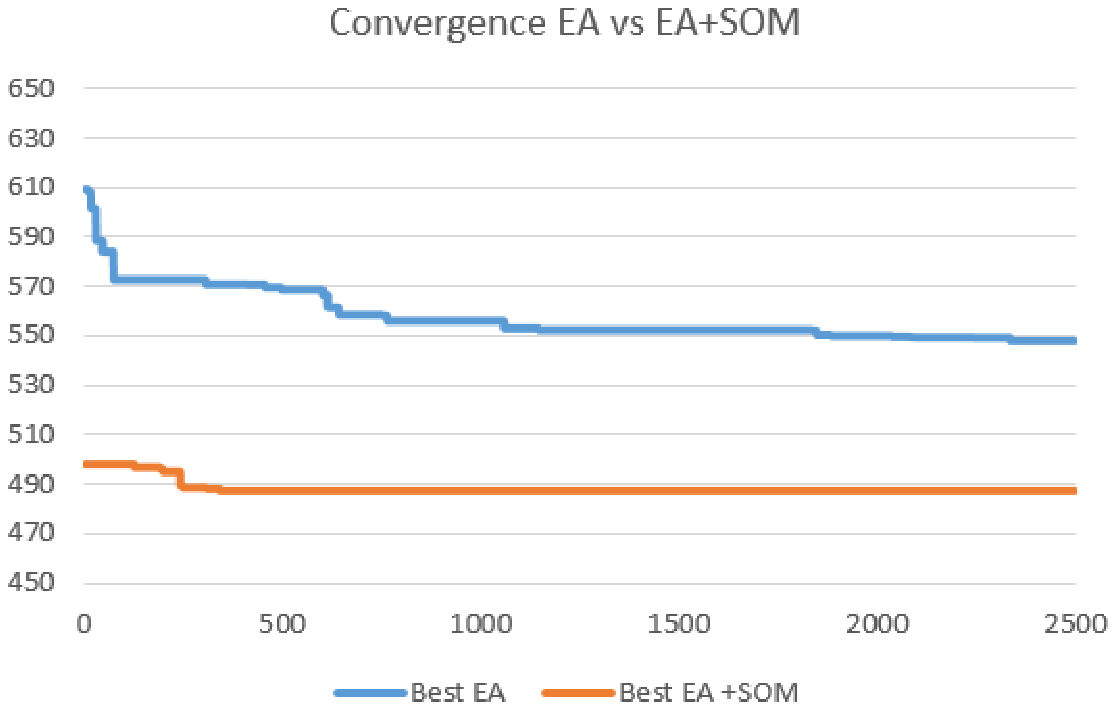}
	\caption{Convergence of the ACO and hybrid ACO algorithms, respectively EA and hybrid EA algorithms for the rat99 problem instance with $m=7$ salesmen}
	\label{fig:conv}
\end{figure}

Figure \ref{fig:conv} plots the evolution of the best solution of the EA and ACO versus the hybrid versions involving SOM on the $rat99$ problem instance. The plot is obtained by averaging over 10 runs for each algorithm and clearly illustrates that the hybrid algorithms have a higher convergence rate than the basic ones.
Both hybrid versions start from better initial solutions than ACO and EA, respectively, and converge to better solutions.

Tables \ref{tab:res1} and \ref{tab:res2} report the mean, the maximum value, the minimum value and the standard deviation of the MinMax objective for each of the considered instances. For each problem instance and each value of $m$, the solution found by CPLEX is given. If CPLEX could not find the optimal value in reasonable time, the upper and the lower bound values are given. The values which are found within the CPLEX interval are highlighted with blue.

\begin{table*}
\centering
\caption{Results obtained by all algorithms for eil51 and berlin52 instances}
\label{tab:res1}
\begin{tabular}{|c|c|c|c|c|c|c|c|c|c|} 
\hline
\multicolumn{2}{|c|}{}               & \multicolumn{4}{c|}{eil51}                    & \multicolumn{4}{c|}{berlin52}          \\ 
\cline{3-10}
\multicolumn{2}{|c|}{}                                 & 2                       & 3      & 5      & 7      & 2       & 3       & 5       & 7        \\ 
\hline
\hline
\multirow{4}[2]{*}{SOM}          & min                    & 236.95                  & 172.81 & 134.94 & 120.45 & 4253.82 & 3475.55 & 2949.61 & 2724.73  \\ 
\cline{2-10}
                              & max                    & 319.19                  & 266.96 & 190.28 & 164.79 & 6574.19 & 5130.73 & 4385.84 & 3950.9   \\ 
\cline{2-10}
                              & avg                    & 278.44                  & 210.25 & 157.68 & 136.84 & 5350.83 & 4197.61 & 3461.93 & 3125.21  \\ 
\cline{2-10}
                              & stdev                  & 15.21                   & 12.85  & 10.85  & 8.21   & 423.89  & 305.68  & 244.02  & 222.93   \\ 
\hline
\hline
\multirow{4}[2]{*}{ACO}          & min                    & 243.81                  & 175.4  & 128.72 & 115.94 & 4366.1  & 3344.66 & 2709.21 & 2448.2   \\ 
\cline{2-10}
                              & max                    & 254.22                  & 185.51 & 138.77 & 122.82 & 4420.87 & 3554.74 & 2753.91 & 2561.54  \\ 
\cline{2-10}
                              & avg                    & 248.76                  & 180.59 & 135.09 & 119.96 & 4388.99 & 3468.9  & 2733.56 & 2510.09  \\ 
\cline{2-10}
                              & stdev                  & 2.85                    & 2.74   & 2.3    & 1.4    & 614.39  & 487.56  & 382.28  & 351.26   \\ 
\hline
\hline
\multirow{4}[2]{*}{SOM-ACO}      & min                    & 238.22                  & 175.4  & 128.56 & 114.34 & 4337.45 & 3352.56 & 2670.82 & 2461.1   \\ 
\cline{2-10}
                              & max                    & 255                     & 185.9  & 140.05 & 123.05 & 4398.15 & 3499.71 & 2760.66 & 2551     \\ 
\cline{2-10}
                              & avg                    & 247.75                  & 180.67 & 134.65 & 119.82 & 4380.72 & 3423.28 & 2719.06 & 2509.07  \\ 
\cline{2-10}
                              & stdev                  & 3.66                    & 2.45   & 2.94   & 2.04   & 613.28  & 480.05  & 380.44  & 350.89   \\ 
\hline
\hline
\multirow{4}[2]{*}{EA}           & min                    & 248.53                  & 190.62 & 134.77 & 116.49 & 4472.83 & 3567.4  & 2694.62 & 2441.39  \\ 
\cline{2-10}
                              & max                    & 299.69                  & 230.64 & 165.64 & 133.34 & 5718.77 & 4254.86 & 3249.35 & 2706.13  \\ 
\cline{2-10}
                              & avg                    & 276.62                  & 208.16 & 151.21 & 123.88 & 5038.33 & 3865.45 & 2853.63 & 2543.73  \\ 
\cline{2-10}
                              & stdev                  & 13.35                   & 8.16   & 7      & 4.04   & 281.72  & 162     & 107.51  & 90.01    \\ 
\hline
\hline
\multirow{4}[2]{*}{SOM-EA}       & min                    & 230.48                  & 163.99 & 124.56 & 112.71 & 4177.42 & \textcolor{blue}{\textbf{3202.66}} & 2627.11 & \textcolor{blue}{\textbf{2440.92}}  \\ 
\cline{2-10}
                              & max                    & 240.15                  & 170.46 & 129.86 & 120.45 & 4188.59 & 3458.08 & 2879.22 & 2706.13  \\ 
\cline{2-10}
                              & avg                    & 235.36                  & 167.85 & 128.57 & 117.48 & 4188.14 & 3368.39 & 2762    & 2498.44  \\ 
\cline{2-10}
                              & stdev                  & 2.47                    & 2.17   & 1.95   & 2.17   & 1.55    & 65.05   & 75.02   & 85       \\ 
\hline
\hline
\multirow{4}[2]{*}{SOM-EA-2opt}       & min                    & \textcolor{red}{\textbf{222.73}}                  & \textcolor{blue}{\textbf{159.57}} & \textcolor{blue}{\textbf{118.54}} & \textcolor{blue}{\textbf{112.07}} & \textcolor{blue}{\textbf{4110.21}} & \textcolor{blue}{\textbf{3073.04}} & 2449.55 & \textcolor{blue}{\textbf{2440.92}}  \\ 

\cline{2-10}
                              & max                    & 236.62                  & 167.06 & 129.86 & 119.95 & 4188.36 & 3314.65 & 2740.96 & 2474.59  \\ 
\cline{2-10}
                              & avg                    & 230.3                   & 164    & 125.64 & 113.75 & 4163.37 & 3193.49 & 2594.13 & 2442.51  \\ 
\cline{2-10}
                              & stdev                  & 2.9                     & 2.97   & 3.15   & 2.12   & 21.46   & 50.52   & 79.74   & 6.66     \\ 
\hline
\hline
\multirow{2}[1]{*}{CPLEX}        & \multirow{2}[1]{*}{value} & \multirow{2}[1]{*}{\textcolor{red}{\textbf{222.73}}} & \textcolor{blue}{\textbf{150.7}}  & \textcolor{blue}{\textbf{96.91}}  & \textcolor{blue}{\textbf{75.91}}  & \textcolor{blue}{\textbf{4049.05}} & \textcolor{blue}{\textbf{2753.63}} & 1671.69 & \textcolor{blue}{\textbf{1272.06}}  \\ 
                              &                        &                         & \textcolor{blue}{\textbf{159.57}} & \textcolor{blue}{\textbf{123.96}} & \textcolor{blue}{\textbf{112.07}} & \textcolor{blue}{\textbf{4110.21}} & \textcolor{blue}{\textbf{3244.37}} & 2441.39 & \textcolor{blue}{\textbf{2440.92}}  \\
\hline
\end{tabular}
\end{table*}

\begin{table*}
\centering
\caption{Results obtained by all algorithms for eil76 and rat99 instances}
\label{tab:res2}
\begin{tabular}{|c|c|c|c|c|c|c|c|c|c|} 
\hline
\multicolumn{2}{|c|}{}               & \multicolumn{4}{c|}{eil76}                         & \multicolumn{4}{c|}{rat99}          \\ 
\cline{3-10}
\multicolumn{2}{|l|}{}                                 & 2                       & 3      & 5      & 7      & 2       & 3      & 5      & 7       \\ 
\hline
\hline
\multirow{4}[2]{*}{SOM}          & min                    & 317.07                  & 242.39 & 181.12 & 159.2  & 843.1   & 678.5  & 566.65 & 512.35  \\ 
\cline{2-10}
                              & max                    & 420.12                  & 345.35 & 257.92 & 211.61 & 1077.62 & 854.69 & 703.98 & 632.37  \\ 
\cline{2-10}
                              & avg                    & 364.02                  & 278.63 & 210.69 & 183.09 & 927.36  & 756.08 & 624.38 & 564.14  \\ 
\cline{2-10}
                              & stdev                  & 19.07                   & 17.21  & 13.4   & 10.82  & 33.41   & 31.91  & 25.07  & 20.98   \\ 
\hline
\hline
\multirow{4}[2]{*}{ACO}          & min                    & 300.43                  & 225.13 & 162.43 & 146.15 & 744.01  & 607.76 & 515.7  & 483.75  \\ 
\cline{2-10}
                              & max                    & 322.81                  & 232.92 & 170.63 & 152.95 & 780.25  & 631.15 & 534.22 & 499.72  \\ 
\cline{2-10}
                              & avg                    & 308.53                  & 224.56 & 163.93 & 146.88 & 767.15  & 620.45 & 525.54 & 492.13  \\ 
\cline{2-10}
                              & stdev                  & 4.66                    & 2.22   & 1.9    & 1.45   & 8.02    & 6.6    & 4.4    & 3.33    \\ 
\hline
\hline
\multirow{4}[2]{*}{SOM-ACO}      & min                    & 306.67                  & 222.03 & 161.41 & 141.92 & 735.43  & 604.69 & 517.07 & 482.72  \\ 
\cline{2-10}
                              & max                    & 317.05                  & 227.79 & 171.93 & 152.12 & 777.64  & 632.58 & 531.83 & 497.87  \\ 
\cline{2-10}
                              & avg                    & 306.23                  & 219.92 & 163.59 & 144.83 & 762.84  & 619.89 & 525.83 & 492.19  \\ 
\cline{2-10}
                              & stdev                  & 2.67                    & 1.34   & 2.44   & 2.01   & 10.3    & 6.19   & 3.45   & 2.7     \\ 
\hline
\hline
\multirow{4}[2]{*}{EA}           & min                    & 342.95                  & 259.21 & 190.5  & 158.78 & 838.37  & 698.72 & 567.85 & 504.01  \\ 
\cline{2-10}
                              & max                    & 388.63                  & 308.07 & 232.24 & 195.57 & 954.92  & 791.01 & 633.04 & 562.95  \\ 
\cline{2-10}
                              & avg                    & 365.72                  & 285.43 & 211.91 & 177.83 & 896.72  & 739.43 & 596.87 & 534.91  \\ 
\cline{2-10}
                              & stdev                  & 10.71                   & 10.8   & 8.91   & 8.1    & 28.01   & 22.84  & 14.22  & 12.18   \\ 
\hline
\hline
\multirow{4}[2]{*}{SOM-EA}       & min                    & 297.55                  & 212.79 & 156.11 & \textcolor{blue}{\textbf{135}}    & \textcolor{blue}{\textbf{709.15}}  & \textcolor{blue}{\textbf{580.79}} & 493.21 & 465.22  \\ 
\cline{2-10}
                              & max                    & 308.67                  & 230.8  & 171.06 & 153.17 & 770.45  & 644.46 & 544.28 & 493.4   \\ 
\cline{2-10}
                              & avg                    & 302.19                  & 220.03 & 164.64 & 143.36 & 736.85  & 612.89 & 519.48 & 482.34  \\ 
\cline{2-10}
                              & stdev                  & 3.46                    & 5.99   & 3.11   & 5.13   & 16.13   & 13.96  & 11.94  & 6.55    \\ 
\hline
\hline
\multirow{4}[2]{*}{SOM-EA-2opt} & min                    & 286.34                  & 205.72 & \textcolor{blue}{\textbf{149.32}} & \textcolor{blue}{\textbf{130.1}}  & \textcolor{blue}{\textbf{680.33}}  & \textcolor{blue}{\textbf{544.13}} & 469.56 & 449.4   \\ 
\cline{2-10}
                              & max                    & 299.96                  & 226.15 & 163.95 & 146.41 & 730.35  & 587.45 & 497.69 & 469.06  \\ 
\cline{2-10}
                              & avg                    & 291.51                  & 211.67 & 156.55 & 137.02 & 703.17  & 564.11 & 483.03 & 458.97  \\ 
\cline{2-10}
                              & stdev                  & 4.08                    & 4.57   & 2.98   & 3.93   & 14.28   & 10.46  & 6.71   & 5.81    \\ 
\hline
\hline
\multirow{2}[1]{*}{CPLEX}        & \multirow{2}[1]{*}{value} & \multirow{2}[1]{*}{280.85} & 186.34 & \textcolor{blue}{\textbf{117.61}} & \textcolor{blue}{\textbf{88.35}}  & \textcolor{blue}{\textbf{620.99}}  & \textcolor{blue}{\textbf{426.25}} & 271.91 & 210.41  \\ 
                              &                        &                         & 197.34 & \textcolor{blue}{\textbf{150.3}}  & \textcolor{blue}{\textbf{139.62}} & \textcolor{blue}{\textbf{728.71}}  & \textcolor{blue}{\textbf{587.17}} & 469.25 & 443.91  \\
\hline
\end{tabular}
\end{table*}

\begin{figure*}[ht]
	\centering
	\includegraphics[width = 9cm]{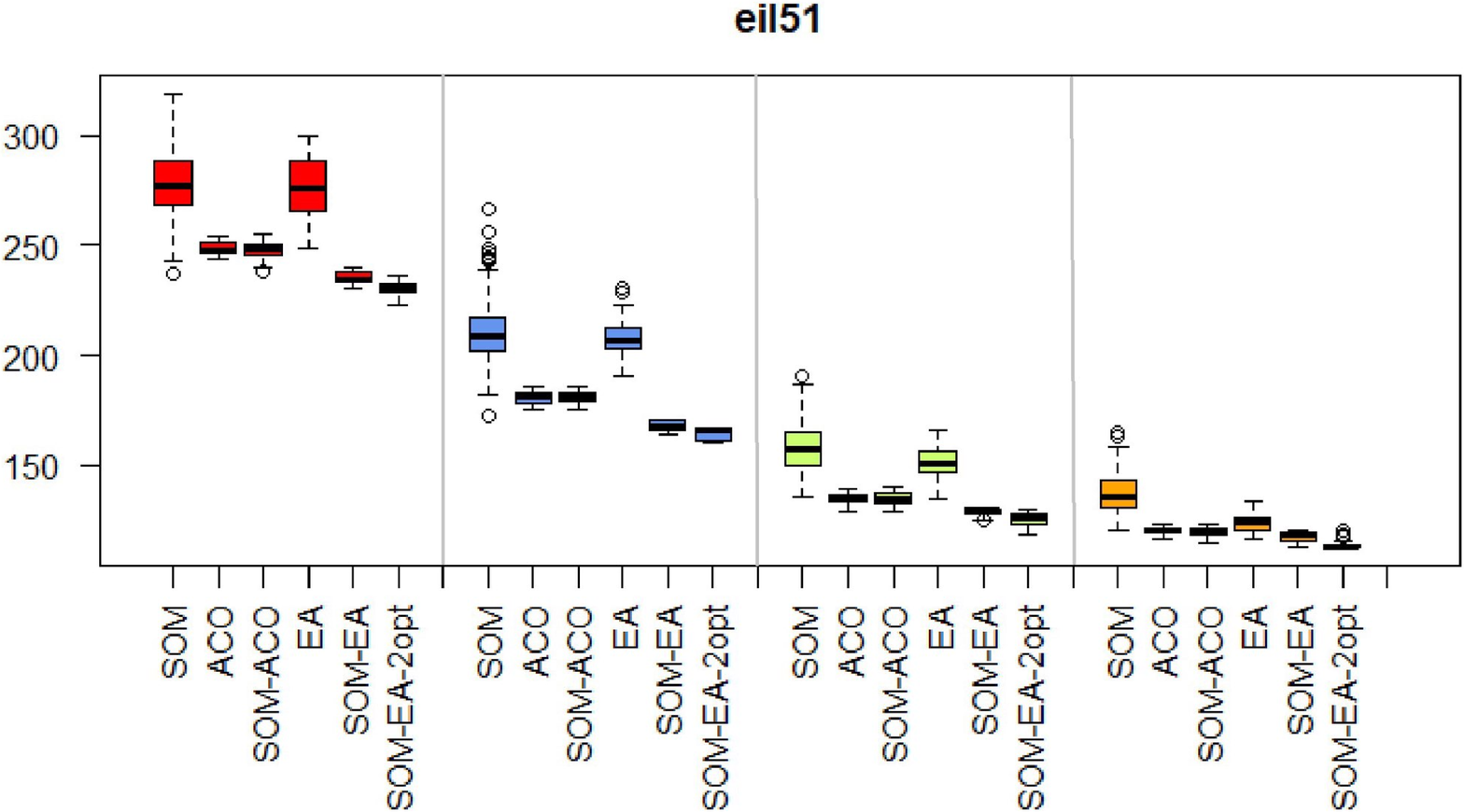}
		\includegraphics[width = 9cm]{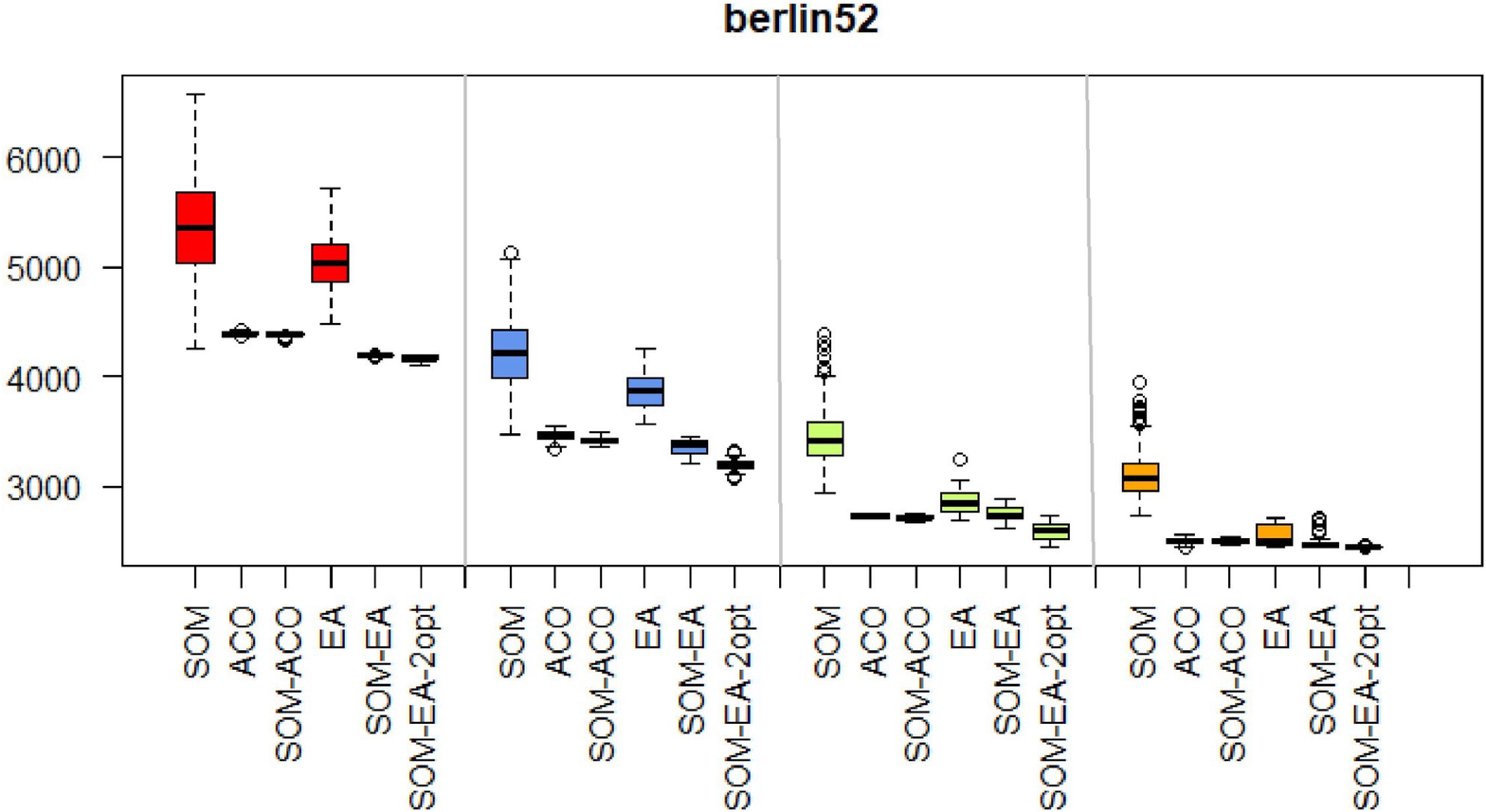}
			\includegraphics[width = 9cm]{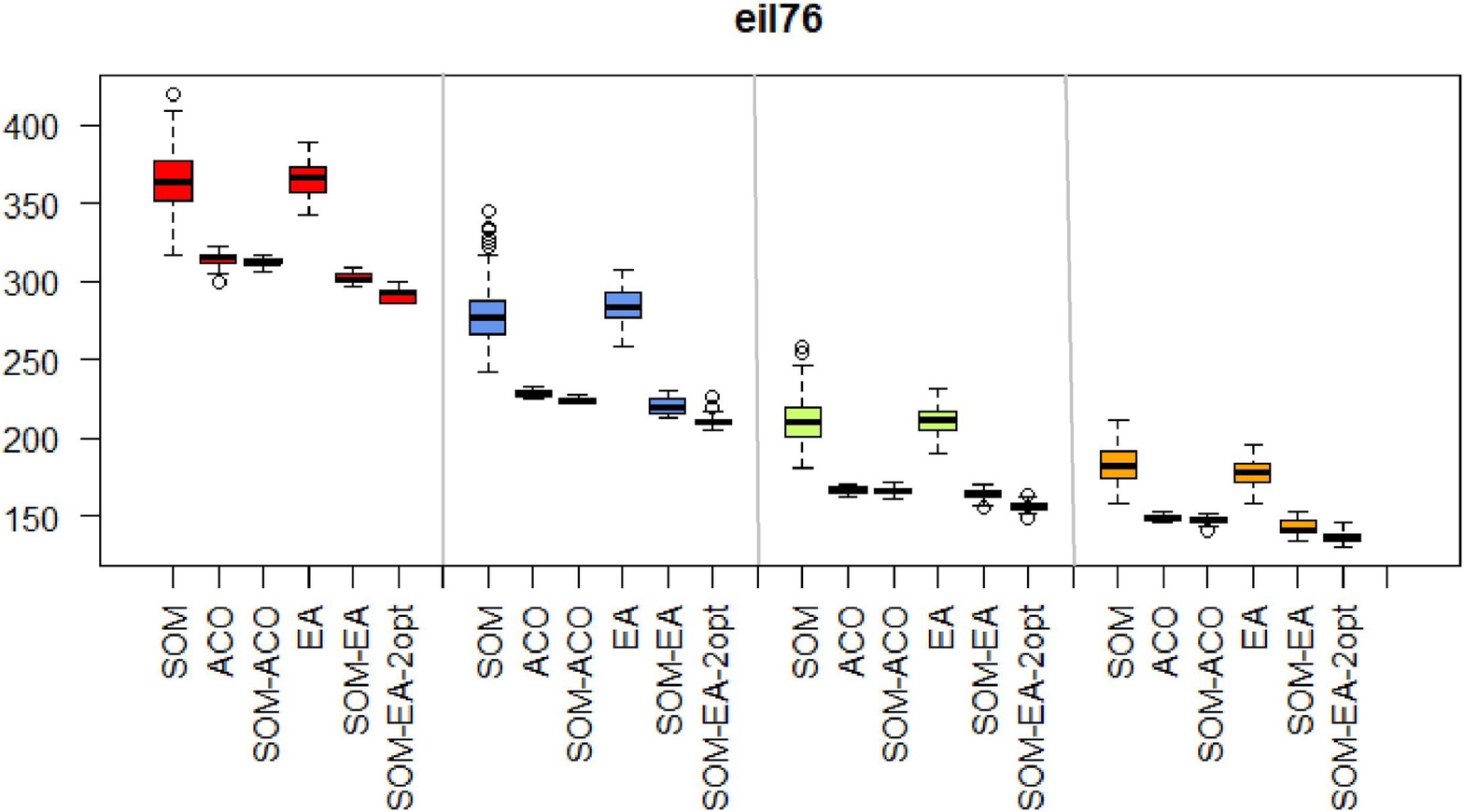}
				\includegraphics[width = 9cm]{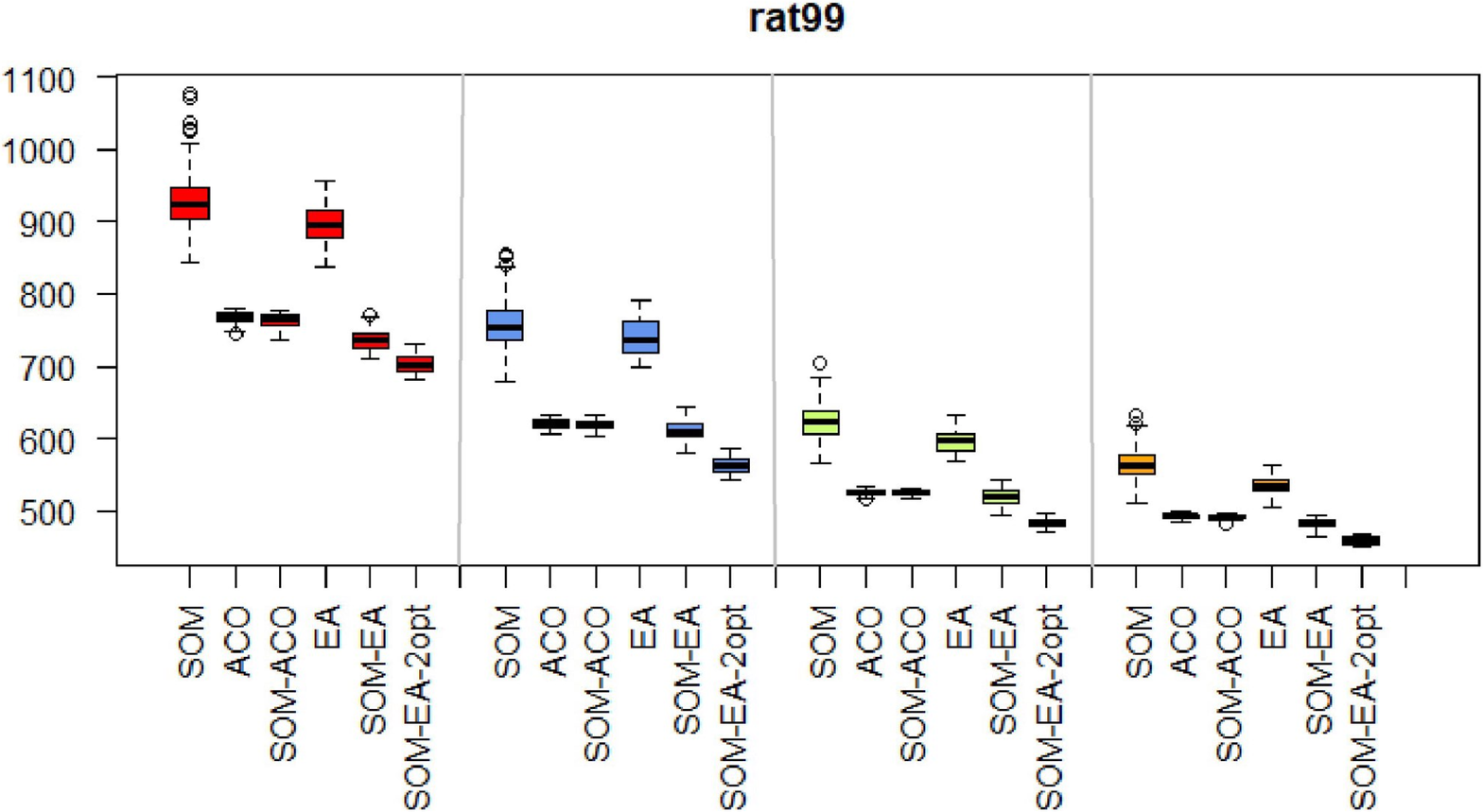}
	\caption{Results for the MinMax objective. Each group corresponds from left to right to different number of salesmen: $m=2,3,5,7$}
	\label{fig:comps}
\end{figure*}

Figure \ref{fig:comps} illustrates the distribution of the MinMax cost obtained by each algorithm. Each chart corresponds to a problem instance and each group in the chart corresponds to a different number of salesmen (2, 3, 5 and 7). 

As seen in Tables \ref{tab:res1} and \ref{tab:res2}, respectively in Figure \ref{fig:comps}, the cost of the longest tour decreases with the increase of $m$, which is the effect of sharing the work across more agents. 

Ant Colony System achieves better results than both SOM and EA. When comparing the hybrid version SOM-ACO with the standalone ant algorithm, a two-sample $t$ test (sample size = 50 runs, significance level 10\%) indicate significant differences in results only for the case of $eil76$ with $m=2,3,7$ salesmen, $berlin52$ with $m=2,3,5$ salesmen and $rat99$ with $m=2$ salesmen.

In the case of the evolutionary algorithm it is evident that the hybridization between EA and SOM improves great over the simple EA in all cases. Regarding the comparative performance of SOM-ACO versus SOM-EA, the latter achieves statistically significant better results on $eil51$ on all the cases, for $eil76$ on $m=2,7$ salesmen, for $berlin52$ on $m=2,3$ salesmen, and for $rat99$ on $m=2,7$ salesmen.

Generally, the hybridization not only improves results but also reduces the variance, which is a desirable feature in practice.

When enhancing SOM-EA with the 2-opt local search heuristic the results also improve significantly. This algorithm found the optimum value for the $eil51$ problem instance with $m=2$ salesmen and outperformed in several cases the approximate solution given by CPLEX within a generous time frame of several days (entries in the table highlighted in blue). 

\section{Conclusions}\label{sec:conclusion}
MinMax Multiple-TSP is addressed by means of three different paradigms: Self Organizing Maps (SOM), Ant Colony Optimization (ACO) and an Evolutionary Algorithm (EA). A hybridization between SOM and the two meta-heuristics show to bring significant improvements, especially in the case of EA. Furthermore, when the 2-opt local-search heuristic is used within the hybrid SOM-EA algorithm, the optimum solution is obtained on some problem instances or solutions better than the ones reported by CPLEX within a time limit of several days are found. The SOM-EA-2opt hybrid is a promising tool for solving MinMax Multiple-TSP and its application on specific Vehicle Routing problems is considered as future work.


\begin{thebibliography}{21}


\bibitem{HAIS2015}
R. Necula, M. Breaban, and M. Raschip, "Performance evaluation of ant colony systems for the single-depot multiple traveling salesman problem," in Proceedings of the 10th International Conference on Hybrid Artificial Intelligence Systems, vol. 9121, pp. 257-268, 2015.

\bibitem{ICTAI2015}
R. Necula, M. Breaban, and M. Raschip, "Tackling the Bi-criteria Facet of Multiple Traveling Salesman Problem with Ant Colony Systems," in Proceedings of the 2015 IEEE 27th International Conference on Tools with Artificial Intelligence, pp. 873-880, 2015.

\bibitem{EVOLVE2015}
R. Necula, M. Raschip, and M. Breaban. "Balancing the Subtours for Multiple TSP Approached with ACS: Clustering-Based Approaches Vs. MinMax Formulation." EVOLVE-A Bridge between Probability, Set Oriented Numerics, and Evolutionary Computation VI. Springer, Cham, 2018. 210-223.
 
\bibitem{Bektas2006} 
Bektas, T. (2006). The multiple traveling salesman problem: an overview of formulations
and solution procedures. Omega, 34(3), 209-219.
 
\bibitem{bektas2013}
T. Bektas. Balancing tour durations in routing a vehicle fleet. In Proceedings of the IEEE Workshop on Computational Intelligence In Production And Logistics Systems,pp. 9-16, 2013.

\bibitem{Franca95}
P.M. Fran\c{c}a, M. Gendreau, G. Laporte, F.M. M\"{u}ller. The m-Traveling Salesman Problem with Minmax Objective. Transportation Science, 29(3), 267-275, 1995.

\bibitem{Chandran06}
N. Chandran, T.T. Narendran, K. Ganesh. A clustering approach to solve the multiple travelling salesmen problem. International Journal of Industrial and Systems Engineering, 1(3), 372-387, 2006.

\bibitem{Wang17}
Y. Wang, Y. Chen, Y. Lin. Memetic algorithm based on sequential variable neighborhood descent for the minmax multiple traveling salesman problem. Computers \& Industrial Engineering, 106, 105-122, 2017.

\bibitem{Vallivaara08}
I. Vallivaara. A team ant colony optimization algorithm for the multiple travelling salesmen problem with minmax objective. In Proceedings of the 27th IASTED International Conference on Modelling, Identification and Control, pp. 387-392, ACTA Press, 2008.

\bibitem{Matsuura14}
T. Matsuura, K. Numata. "Solving Min-Max Multiple Traveling Salesman Problems by Chaotic Neural Network". International Symposium on Nonlinear Theory and its Applications, 2014.

\bibitem{Modares99}
A. Modares, S. Somhom, T. Enkawa. "A self‐organizing neural network approach for multiple traveling salesman and vehicle routing problems". International Transactions in Operational Research, 6(6), 591-606, 1999.

\bibitem{Carter06}
A.E. Carter, C.T. Ragsdale. A new approach to solving the multiple traveling salesperson problem using evolutionary algorithms. European Journal of Operational Research, 175(1), 246-257, 2006.

\bibitem{Kiraly11}
A. Kir\'aly, J. Abonyi. Optimization of multiple traveling salesmen problem by a novel representation based evolutionary algorithm. In Intelligent Computational Optimization in Engineering, pp. 241-269, Springer, 2011.


\bibitem{SOM88a}
B. Angéniol, G. de La Croix Vaubois, J.-y. Le Texier. "Self-organizing feature maps and the travelling salesman problem". Neural Networks. 1988;1(4):289–293, 1998.

\bibitem{SOM88b}
J.-C.Fort. "Solving a combinatorial problem via self-organizing process: an application of the Kohonen algorithm to the traveling salesman problem." Biological Cybernetics. 1988;59(1):33–40, 1988.  

\bibitem{Faigl16}
J. Faigl. An application of self-organizing map for multirobot multigoal path planning with minmax objective. Computational intelligence and neuroscience, vol. 2016 (2016): 2720630.

\bibitem{Croes58}
G.A. Croes. A Method for Solving Traveling-Salesman Problems. Operation Research, 6, 791-812, 1958.

\bibitem{gavrilut1}
D. Gavrilut, R. Benchea, C. Vatamanu. Optimized zero false positives perceptron training for malware detection. In 2012 14th International Symposium on Symbolic and Numeric Algorithms for Scientific Computing (pp. 247-253), 2012.

\bibitem{gavrilut2}
C Vatamanu, D Gavriluţ, RM Benchea. Building a practical and reliable classifier for malware detection.Journal of Computer Virology and Hacking Techniques, 2013.

\end{thebibliography}
\end{document}